\documentclass[10pt,twocolumn,letterpaper]{article}
\usepackage[pagenumbers]{cvpr} 

\usepackage{graphicx}
\usepackage{amsmath}
\usepackage{amssymb}
\usepackage{booktabs}
\usepackage{soul}
\usepackage[toc,page,header]{appendix}
\usepackage{minitoc}

\usepackage[pagebackref,breaklinks,colorlinks]{hyperref}

\usepackage[table,x11names]{xcolor}
\definecolor{aliceblue}{rgb}{0.94, 0.97, 1.0}
\definecolor{mistyrose}{rgb}{1.0, 0.89, 0.88}
\DeclareRobustCommand{\MA}[1]{{\sethlcolor{aliceblue}\hl{#1}}}

\usepackage[capitalize]{cleveref}
\crefname{section}{Sec.}{Secs.}
\Crefname{section}{Section}{Sections}
\Crefname{table}{Table}{Tables}
\crefname{table}{Tab.}{Tabs.}

\usepackage{overpic}
\usepackage{enumitem} 
\usepackage{overpic} 
\usepackage{color}

\definecolor{turquoise}{cmyk}{0.65,0,0.1,0.3}
\definecolor{purple}{rgb}{0.65,0,0.65}
\definecolor{dark_green}{rgb}{0, 0.5, 0}
\definecolor{orange}{rgb}{0.8, 0.6, 0.2}
\definecolor{red}{rgb}{0.8, 0.2, 0.2}
\definecolor{darkred}{rgb}{0.6, 0.1, 0.05}
\definecolor{blueish}{rgb}{0.0, 0.3, .6}
\definecolor{light_gray}{rgb}{0.7, 0.7, .7}
\definecolor{pink}{rgb}{1, 0, 1}
\definecolor{greyblue}{rgb}{0.25, 0.25, 1}

\usepackage{blindtext}

\renewcommand{\paragraph}[1]{\vspace{1em}\noindent\textbf{#1}.}
\begin{document}
\doparttoc 
\faketableofcontents 

\title{Learning Audio-Video Modalities from Image Captions}

\author{Arsha Nagrani \quad Paul Hongsuck Seo \quad Bryan Seybold \\
Anja Hauth \quad Santiago Manen \quad Chen Sun \quad Cordelia Schmid \\
Google Research\\
{\tt\small \{anagrani,phseo,seybold,ahauth,smanen,chensun,cordelias\}@google.com}
}
\maketitle
\begin{abstract}
A major challenge in text-video and text-audio retrieval is the lack of large-scale training data. This is unlike image-captioning, where datasets are in the order of millions of samples. To close this gap we propose a new video mining pipeline which involves transferring captions from image captioning datasets to video clips with no additional manual effort. Using this pipeline, we create a new large-scale, weakly labelled audio-video captioning dataset consisting of millions of paired clips and captions. We show that training a multimodal transformed based model on this data achieves competitive performance on video retrieval and video captioning, matching or even outperforming HowTo100M pretraining with 20x fewer clips. We also show that our mined clips are suitable for text-audio pretraining, and achieve state of the art results for the task of audio retrieval. 
\end{abstract}
\section{Introduction}
\label{sec:intro}
A key facet of human intelligence is the ability to effortlessly connect the visual and auditory world to natural language concepts. Bridging the gap between human perception (visual, auditory and tactile) and communication (via language) is hence becoming an increasingly important goal for artificial agents, enabling tasks
such as text-to-visual retrieval~\cite{wang2016learning,patrick2020support,bain2021frozen}, image and video captioning~\cite{vinyals2016show,you2016image,krishna2017dense}, and visual question answering~\cite{antol2015vqa,lei2018tvqa}. In the image domain in particular, this has lead to an explosion of large scale image datasets with natural language descriptions~\cite{lin2014microsoft,krishna2017visual,sharma2018conceptual,changpinyo2021conceptual}. 
\begin{figure}[t]
\begin{center}
\includegraphics[width=\linewidth]{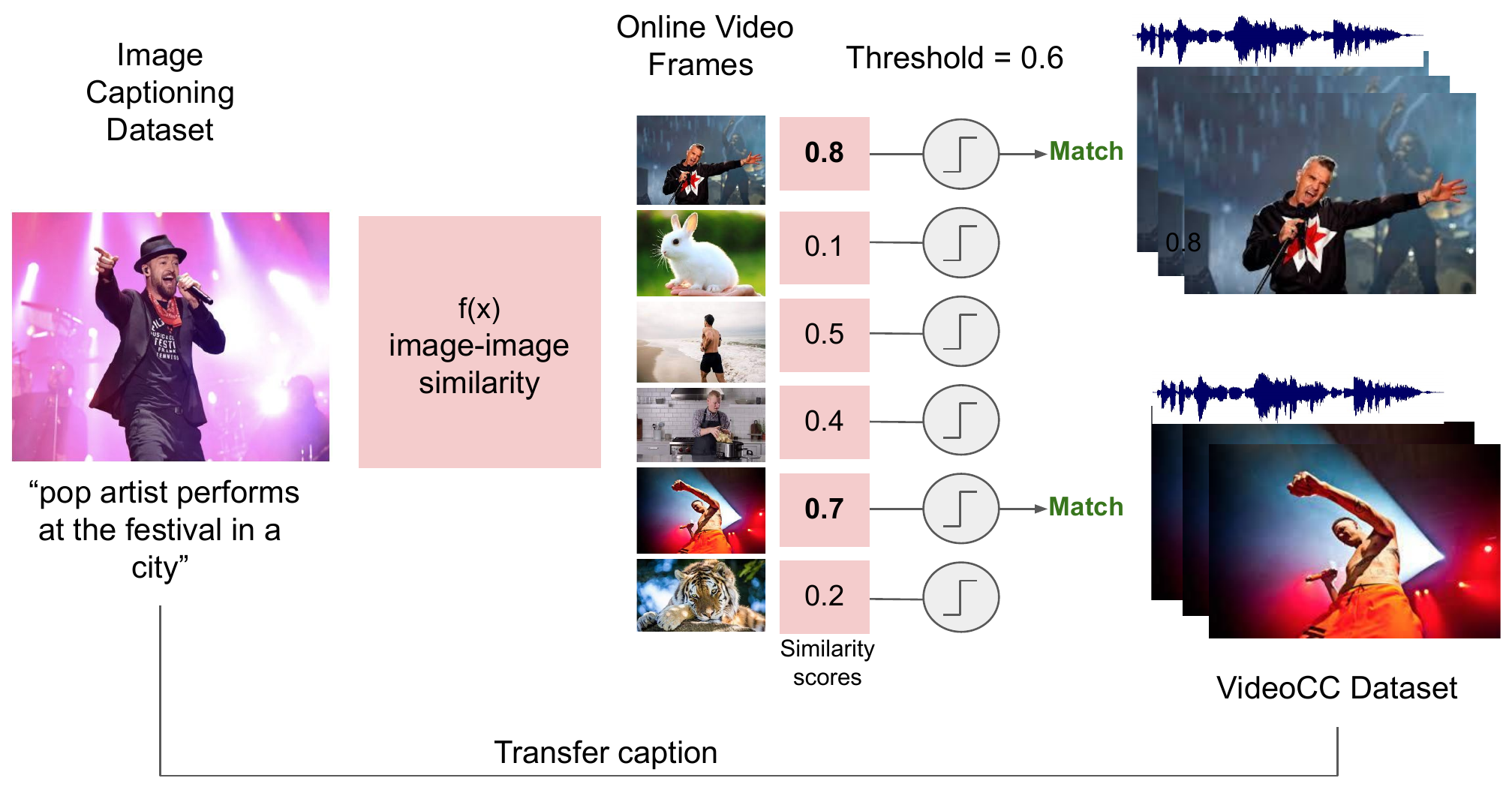}
\end{center}
\caption{
\textbf{Mining Audio-video clips automatically.}
We use the images in image captioning datasets as `seed' frames to mine related audio-visual clips. For each seed image-caption pair in a dataset, we find frames in videos with high similarity scores to the seed image. We then extract short video clips around the matching frames and transfer the caption to those clips. This gives us free captioning supervision for video and audio clips.
}
\label{fig:teaser}
\end{figure}

 In the video and audio domains, however, recent research seems to either be directed at modelling, for example in developing new architectures (eg.\ multimodal transformers \cite{gabeur2020multi,seo2021look,bain2021frozen}), or new training objectives (eg.\ those that can deal with misaligned~\cite{miech20endtoend} or overly specialised~\cite{patrick2021supportset} inputs). Often over-looked is the underlying data used to train and evaluate models. Annotating videos manually with clean and diverse captions is often subjective, painstaking and expensive. This means that most video-captioning datasets (eg.\ MSR-VTT~\cite{xu2016msr}, LSMDC~\cite{rohrbach2017movie}, CMD~\cite{bain2020condensed}, ActivityNet~\cite{krishna2017dense} etc.) are small in size (order of magnitude 100K). Audio captioning datasets such as AudioCaps~\cite{kim2019audiocaps} and Clotho~\cite{drossos2020clotho}, are even smaller.

Given the well-known benefits of pretraining, the community has been forced to look at creative but weak forms of supervision, such as hashtags~\cite{ghadiyaram2019large}, titles and descriptions~\cite{stroud2020learning}, or Automatic Speech Recognition (ASR) in instructional videos~\cite{miech2019howto100m}. The de facto standard for video-language pretraining~\cite{li2020hero,amrani2020noise,luo2020univilm,rouditchenko2020avlnet,gabeur2020multi,patrick2020support} has become the large HowTo100M~\cite{miech2019howto100m} dataset, pretraining on which gives a significant boost over training from scratch. The pitfalls of using ASR however are well known; (i) there is noise in imperfect ASR transcription, (ii) continuous narration may consist of incomplete or grammatically incorrect sentences, (iii) the domain is often limited to instructional videos to increase relevance between speech and video content and finally, and (iv) ASR may not be temporally aligned with the video, or indeed may not refer to the video at all~\cite{miech2019howto100m}. Combined, this necessitates a huge amount of training data for good performance (100s of millions of samples), and consequently, a lot of compute. 

Image annotation, on the other hand, is cheaper than video and easier to obtain from web pages~\cite{sharma2018conceptual,changpinyo2021conceptual}, and large-scale image-text pretrained models such as CLIP~\cite{radford2021learning} are available online. This has led to concurrent works~\cite{luo2021clip4clip,fang2021clip2video} using image-text models for video-text tasks. While this is a valuable idea, using such models beyond weight initialization requires some additional complexity. If we treat videos as a bag of sparse frames~\cite{lei2021less}, we lose all the benefits of video (modalities like audio and the chance to model low-level temporal information directly from the frames) or require complicated distillation procedures from image to video models~\cite{girdhar2019distinit}.
Hence we believe there is still a necessity for large-scale \textit{video}-text datasets. 

Is there another way to leverage all the existing effort that has gone into image-captioning datasets? 
We propose a solution in the form of a new video mining method based on \textit{cross-modal transfer}, where we use images from image captioning datasets as seeds to find similar clips in videos online (Fig. \ref{fig:teaser}). We then transfer the image captions directly to these clips, obtaining weak, albeit free video and audio captioning supervision in the process. This can also provide us with motion and audio supervision -- for example, sometimes human-generated captions for images infer other modalities, eg.\ the caption `Person throws a pitch during a game against university' from the CC3M dataset~\cite{sharma2018conceptual} was written for a single, still image, but is actually describing motion that would occur in a video. Similarly, the caption `A person singing a song', is also inferring a potential audio track. We note that like HowTo100M, our dataset curation is entirely automatic, and requires no manual input at all. However, as we show in Sec. \ref{sec:mining-pipeline}, our mined data samples are more diverse than HowTo100M, are matched to better-formed captions compared to ASR, and are likely to contain at least one frame that is aligned with the text caption.

In doing so we make the following contributions: (i) We propose a new, scalable video-mining pipeline which transfers captioning supervision from image datasets to video and audio. (ii) We use this pipeline to mine paired video and captions, using the Conceptual Captions3M~\cite{sharma2018conceptual} image dataset as a seed dataset. Our resulting dataset VideoCC3M consists of millions of weakly paired clips with text captions and will be released publicly.
(iii) We propose a new audio-visual transformer model for the task of video retrieval, which when trained on this weakly paired data performs on par with or better than models pre-trained on HowTo100M for video retrieval and captioning, with 20x fewer clips and 100x fewer text sentences. In particular, we show a large performance boost in the zero-shot setting. (iv) Finally, we also show that our audio-visual transformer model seamlessly transfers to \textit{text-audio} retrieval~\cite{oncescu2021audio} benchmarks as well, achieving state of the art results on the AudioCaps~\cite{kim2019audiocaps} and Clotho~\cite{drossos2020clotho} datasets.

\section{Related work}
\label{sec:related}

\noindent\textbf{Cross-modal supervision:}
Our key idea is to use labelled data in one modality (images) to aid learning in another modality (videos). A popular method for cross-modal transfer is knowledge \textit{distillation}~\cite{hinton2015distilling}, which has shown great success for transferring supervision from RGB to depth~\cite{Gupta_2016_CVPR}, or faces to speech~\cite{albanie2018emotion}. Another line of work enhances unimodal models via multimodal regularisations~\cite{abavisani2019improving,aguilar2019multimodal}.
Ours is a related but tangential idea which involves mining new data and assigning labels to it (similar to video clips mined for action recognition using speech by~\cite{nagrani2020speech2action,gao2020listen}). This is particularly useful when there are large labelled datasets in one modality
(here text-image retrieval~\cite{lin2014microsoft,krishna2017visual,sharma2018conceptual}), but it is more challenging to obtain for a similar task in another modality (text-audio~\cite{oncescu2021audio} or text-video~\cite{xu2016msr,anne2017localizing,krishna2017dense,rohrbach2017movie,zhou2018towards,bain2020condensed} retrieval).\\
\noindent\textbf{Text supervision for video:}
Existing manually annotated video captioning datasets~\cite{xu2016msr,zhou2018towards,huang2020multimodal} are orders of magnitude smaller than classification datasets~\cite{kay2017kinetics}. This has led to a number of creative ideas for sourcing weakly paired text and video data. \cite{sun2015domain} use web images queried with sports activities to create temporal annotations for videos. \cite{ghadiyaram2019large} and ~\cite{li2020learning} use hashtags and titles for supervision respectively, but only to learn a better video encoder. In the movie domain, ~\cite{bain2020condensed} uses YouTube descriptions for movie clips while 
\cite{rohrbach2017movie} uses audio description (AD) from movies. 
The recently released WebVid2M dataset~\cite{bain2021frozen} comprises manually annotated captions, but given the monetary incentive on stock sites, they often contain added metatags appended, and most lack audio. Another valuable recent dataset is Spoken Moments in Time~\cite{monfort2021spoken}, however this was created with significant manual effort. The largest video-text dataset by far is HowTo100M~\cite{miech2019howto100m} generated from ASR in instructional videos; however, this data is particularly noisy, as discussed in the introduction. \\
\noindent\textbf{Text supervision for audio:} 
Textual supervision for audio is even scarcer than it is for video. Early works perform text-audio retrieval using single word audio tags as queries~\cite{chechik2008large}, or class labels as text labels~\cite{elizalde2019cross}. Even earlier,~\cite{slaney2002semantic} linked text to audio but only using 215 animal sounds from the BBC Sound Effects Library. Unlike these works, we study unconstrained caption-like descriptions as queries. While small, manually annotated datasets such as  AudioCaps~\cite{kim2019audiocaps} and Clotho~\cite{drossos2020clotho} do exist (and have been repurposed by~\cite{oncescu2021audio,koepke2021audio} for audio-text retrieval), large-scale pretraining data for text-audio tasks is not available. Note that extracting audio from existing video-text datasets is difficult: WebVid videos largely do not have audio, and HowTo100M captions are derived from the audio (training a model to predict HowTo100M captions from the audio might simply be learning how to do ASR). Hence we explore the link between audio and text transferred via image similarity to videos that all have audio, and show this improves text-audio retrieval. As far as we are aware, we are the first work to pre-train the same model for both \textit{visual-focused} datasets such as MSR-VTT and \textit{audio-focused} datasets such as AudioCaps and Clotho.

\section{Text-video data} \label{sec:mining-pipeline}
In this section we describe our automatic mining pipeline for obtaining video clips paired with captions. We then train text-video and text-audio models (described in Sec. \ref{sec:model}) on this weakly paired data for 3 tasks, video retrieval, video captioning and audio retrieval.

\subsection{Mining pipeline}
The core idea of our mining pipeline is to start with an image captioning dataset, and for each image-caption pair in a dataset, find frames in videos similar to the image. We then extract short video clips around the matching frames and transfer the caption to those clips. In detail, the steps are as follows: \\
\begin{figure*}[t]
\begin{center}
\includegraphics[width=0.9\linewidth]{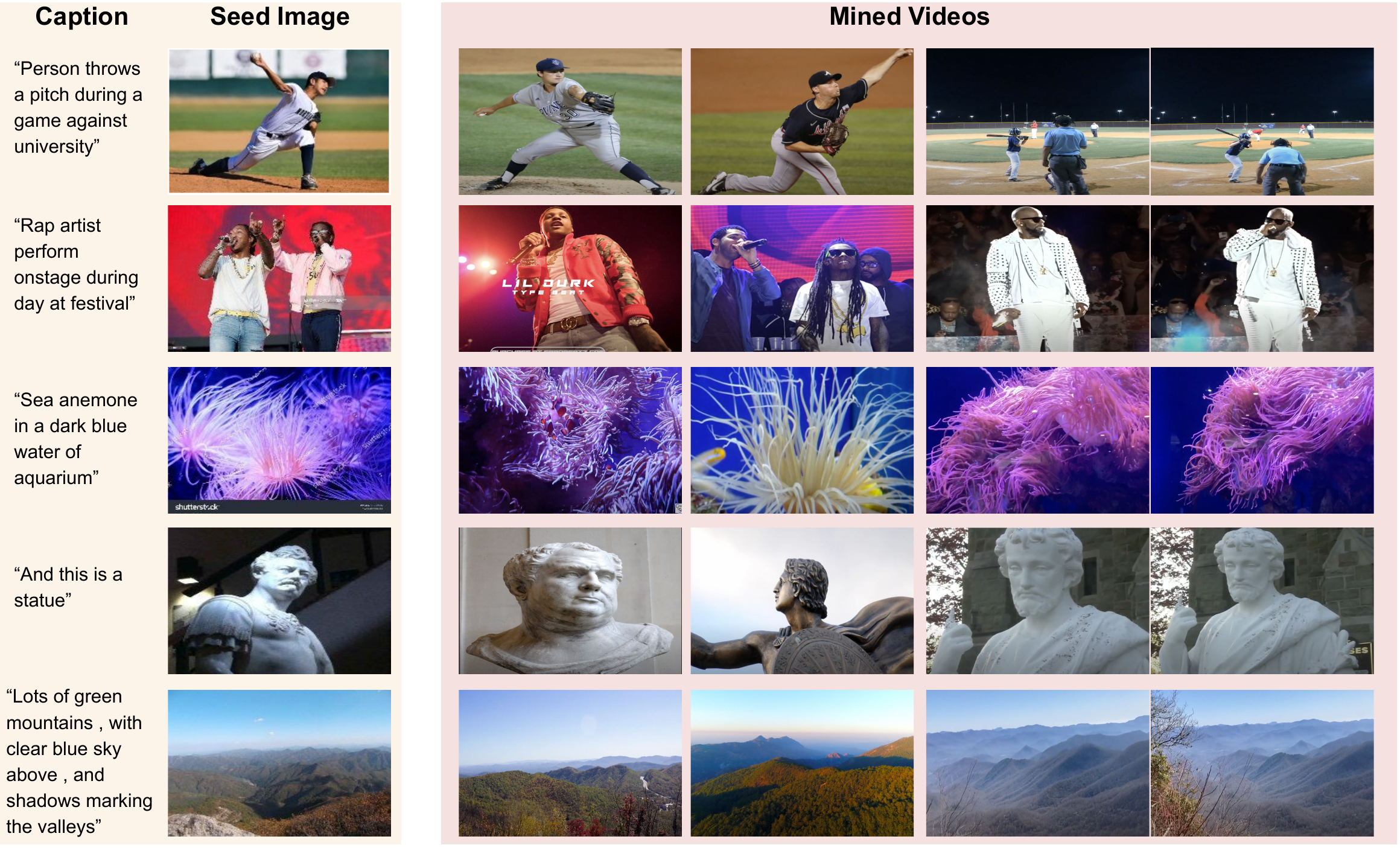}
\end{center}
\caption{
\textbf{Examples of clips with captions that are mined automatically.} For each seed image, we show 3 `matched' clips obtained using our automatic video mining method. For the first 2 clips, we show only a single frame, but for the third clip we present 2 frames to show motion, either of the subjects in the video (first 3 rows) or small camera motion (last 2 rows). Note the diversity in the mined clips, for example the different pitching poses and angles (first row) and the different types of statues (fourth row). Clips in the second row also contain audio relevant to the caption. Note frames may have been cropped and resized for ease of visualisation. More results are provided in the appendix.
}
\label{fig:mining-examples}
\end{figure*}
\noindent\textbf{1. Identify seed images:} We begin by selecting an image-captioning dataset. The images in this dataset are henceforth referred to as `seed' images ($x_\text{seed}$).

\noindent\textbf{2. Feature Extraction:} We then calculate a visual feature vector $f(x_\text{seed})$ for each seed image. Given our primary goal is to mine semantically similar images, we extract features using a deep model trained for image retrieval, the Graph-Regularized Image Semantic
Embedding (Graph-RISE) model~\cite{juan2019graph}. We then extract the same visual features $f(x_v)$ for the frames $x_v$ of a large corpus of videos online. Because visual information in videos is strongly correlated over time, we can extract features at a reduced rate (1fps) relative to the original video frame rate for efficiency.

\noindent\textbf{3. Identify matches:} Next, we calculate the dot product similarity between the feature vectors for each seed image in the caption data set and those for each video frame obtained from the video corpus. Pairs with a similarity above a threshold $\tau$ are deemed `matches'. For each seed image, we keep the top 10 matches. For these top 10, we transfer the caption from the image to a short video clip extracted at a temporal span $t$ around the matched image frame, and add it to our dataset. 
In Sec.~\ref{sec:mining-ablations}, we provide brief ablations on the values of $t$ and the threshold $\tau$.
\subsection{Video-Conceptual-Captions (VideoCC)}
We ran our mining pipeline with the image captioning dataset - Conceptual Captions 3M~\cite{sharma2018conceptual} (CC3M). We only use the images in the dataset which are still publicly available online, which gives us 1.25 image-caption pairs. We apply our pipeline to online videos. We filter videos for viewcount $>$ 1000, length $<$ 20 minutes, uploaded within the last 10 years, but at least 90 days ago, and filter using content-appropriateness signals to get 150M videos.
This gives us 10.3M clip-text pairs with 6.3M video clips (total 17.5K hours of video) and 970K unique captions. We call the resulting dataset VideoCC3M.

\begin{figure}[h]
\begin{center}
\includegraphics[width=0.94\linewidth]{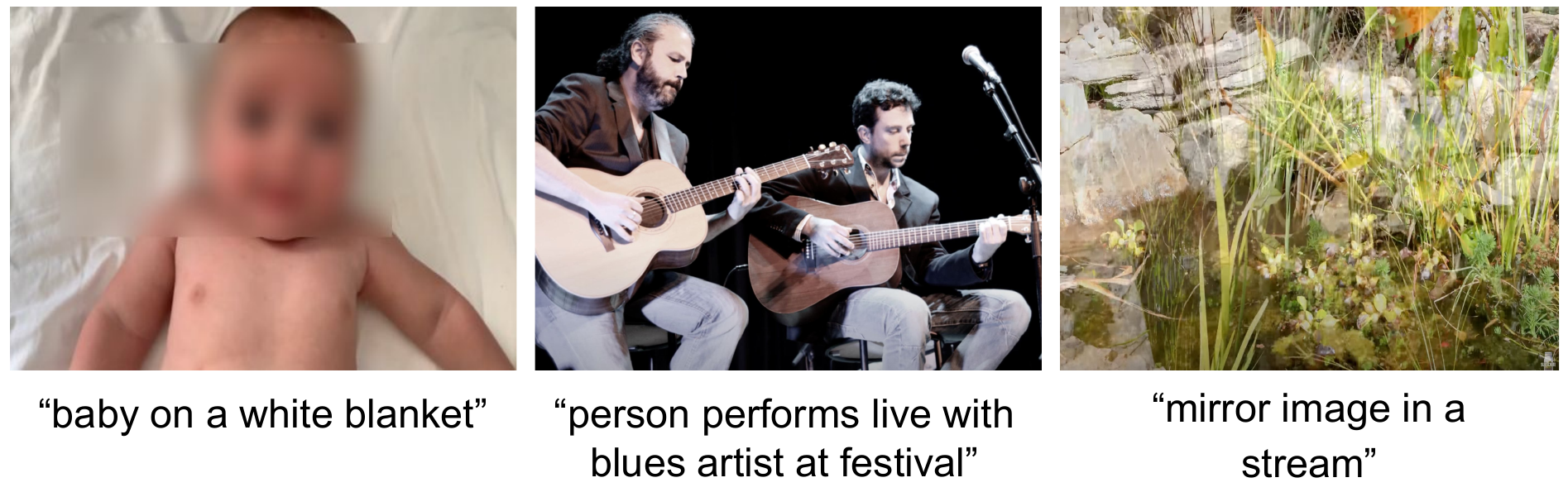}
\end{center}
\vspace{-1em}
\caption{
\textbf{Examples from VideoCC3M of automatically mined clips with relevant audio to the caption.} We show a single relevant frame from each clip as a proxy for visualising the audio. The accompanying audio contains (left to right) the sounds of a baby gurgling, music and water flowing sounds. The left image is intentionally blurred.
}
\label{fig:audio-examples}
\end{figure}

We also run our pipeline on a more recently released extension, called Conceptual Captions 12M~\cite{changpinyo2021conceptual} (CC12M). Note that while CC3M consists of higher quality captions~\cite{sharma2018conceptual}, CC12M was created by relaxing the data collection pipeline used in CC3M, and hence the captions are far noisier. Results on this dataset are provided in the appendix. 
 Some examples of the matched video frames to captions for VideoCC3M are provided in Figure~\ref{fig:mining-examples}. 
\begin{figure}[h]
\begin{center}
\includegraphics[width=0.95\linewidth]{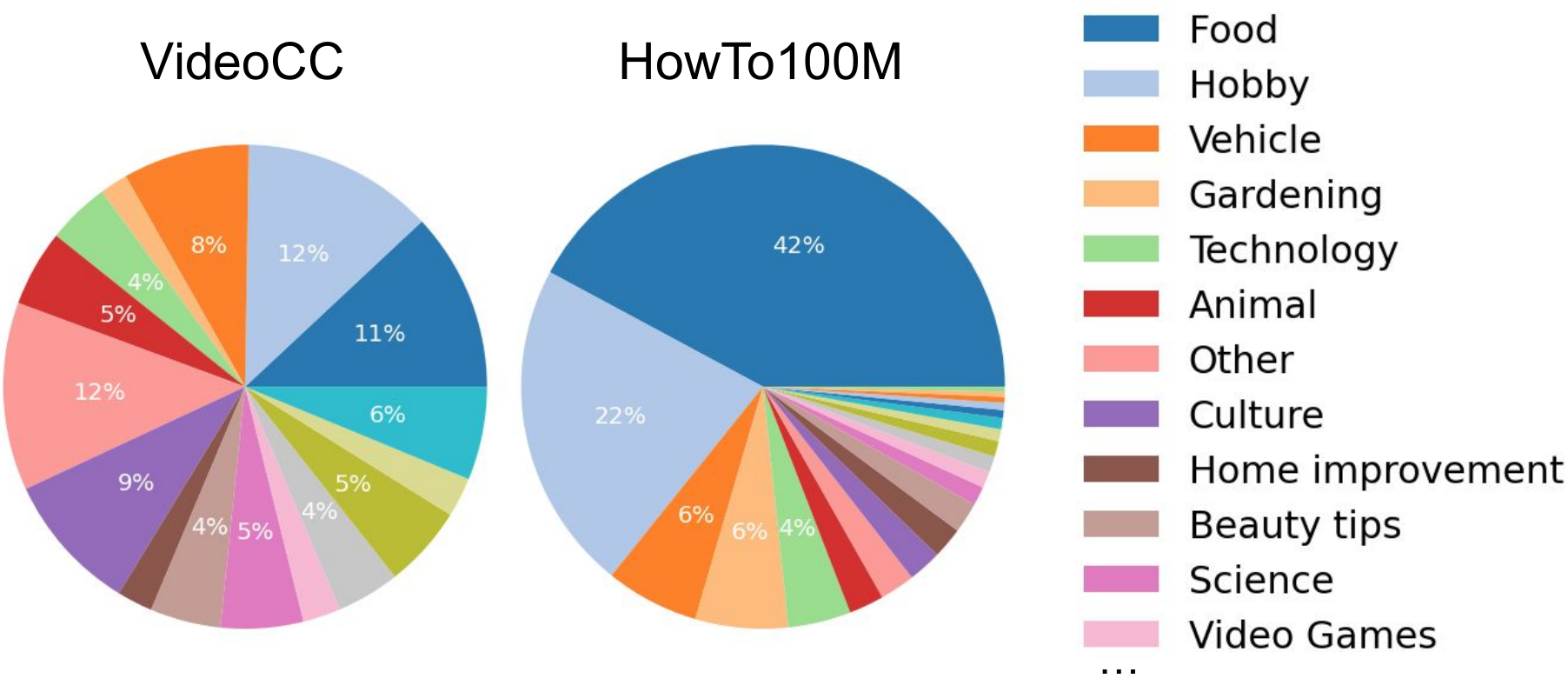}
\end{center}
\caption{
\textbf{Domains in VideoCC3M vs HowTo100M.} VideoCC3M has a more diverse and balanced range of domains, `Other' here includes a variety of content such as music videos, sports, politics, vlogs and so on. Note how almost half of HowTo100M videos are food-related (cooking videos). More details are provided in the appendix.}
\label{fig:domains}
\end{figure}
The mined video clips have the following properties: \\ 
\noindent\textbf{(i) Diversity:} Note that because VideoCC3M is mined from a general corpus of videos online (unlike HowTo100M, which is restricted to instructional videos), our dataset is more balanced. A more comprehensive bar chart is provided in the appendix. Some of the ‘Other’ categories are technology, team sports, family, medicine, beauty, history, religion, gardening, music, politics -- while HowTo100M videos are largely dominated by the ‘Food’ and ‘Hobby’ domains (almost half are `cooking videos'). This is unsurprising given that HowTo100M is limited to instructional videos.\\
\noindent\textbf{(ii) Alignment:} We mine frames that have high visual similarity to the seed image. If this seed has a relevant caption (largely the case for the high quality CC3M dataset), it is likely that at least one frame in the mined clip is aligned with the caption. A manual check of a small subset of clips found this to be the case in 91\% (see suppl). This is a stricter constraint than ASR based datasets, which have occasional misalignment between speech and frames. \\ 
\noindent\textbf{(iii) Caption Style:} The quality of the captions is transferred directly from the seed dataset. Most of the captions in CC3M are fully formed, grammatically correct sentences, unlike the distribution of sentences obtained from ASR. Each caption is matched to a mean of 10.6 clips, with some captions matched to more than 10 clips. This is possible because, while we limit the clip mining to 10 clips per seed image, the original CC3M dataset has multiple seed images with the same caption, eg `an image of digital art', leading to more than 10 mined clips for these captions.\footnote{Full distribution of clips per caption in VideoCC3M is provided in suppl. material.} Having multiple pairs from the same set of captions and video clips also helps ensure that learnt video and text representations are not overly-specialised to individual samples (which can be a problem for existing datasets, as noted by~\cite{patrick2021supportset}).\\
\noindent\textbf{Cross-modal transfer from the image domain} 
Interestingly, this mining method provides us with \textit{captioning} supervision for modalities such as video and audio that are difficult to annotate. Note that we use two existing sources of image supervision, the first is the seed image captioning dataset, and the second is the image similarity model $f(\cdot)$ which we use to mine related frames.
This is not the same as simply applying a text-image model (even though that is a complementary idea) to different frames in a video for text-video retrieval. For example, our method provides some valuable supervision for new clips with motion (see the last column of retrieved clips in Fig. \ref{fig:mining-examples}, first two rows).  Many image captions in CC3M describe actions/motion, eg.~\textit{human-human interactions} (`baby smiling down at dad while being thrown in the air'), \textit{interactions with objects/body parts} (`person shaves hair on neck', `rugby player fields a punt'), \textit{movement in an environment} (`elderly couple walking on a deserted beach').\footnote{We find that interestingly, 83\% of the 7.9K verbs (extracted using spacy package) in MSR-VTT (video annotated dataset), are present in CC3M.} Our mining method, since it retrieves videos, can actually find examples of these described motions. We also obtain some free supervision for the audio stream (Fig. \ref{fig:mining-examples}, second row and Fig. \ref{fig:audio-examples}). These weakly labelled audio samples can be used for pretraining text-audio models, as we show in the results. \\

\subsubsection{Data mining ablations}\label{sec:mining-ablations}
\begin{figure}[h]
\begin{center}
\includegraphics[width=0.96\linewidth]{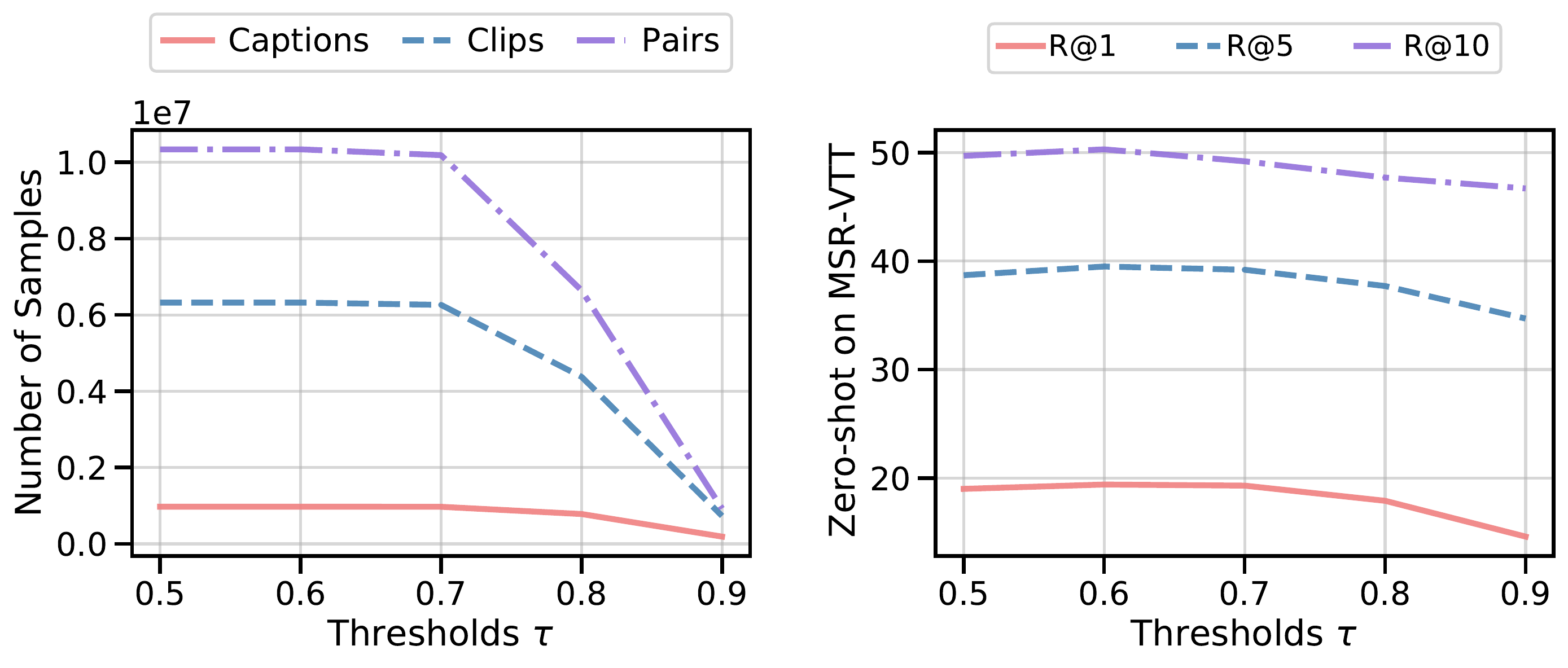}
\end{center}
\caption{
\textbf{Effect of match threshold $\tau$} on mining statistics (left) and zero-shot performance on MSR-VTT (right). Increasing the threshold beyond 0.6 decreases the size of the dataset, which leads to a corresponding performance drop on zero-shot retrieval. We use an optimal match threshold of $0.6$. }
\label{fig:threshold}
\end{figure}

In this section, we ablate the value of the time span $t$ and threshold $\tau$. We use zero-shot performance on the MSR-VTT test set (this protocol is described in Sec. \ref{sec:retrieval}) to test these ablations. \\
\noindent\textbf{Time span $t$:} We try extracting different length clip segments $t$ between 5 and 30 seconds, and found that performance increases up until 10 seconds, but decreases after that (results and discussion in the suppl. material). Hence we extract $10$ second clips for our dataset. \\
\noindent\textbf{Match threshold $\tau$:} We experiment with different match thresholds $\tau$ for the similarity in the range $\{0.5,0.6,0.7,0.8.0.9\}$ and present the effect of this on mining statistics in Figure \ref{fig:threshold}. The higher the match threshold, the stricter the similarity requirement on the matched frames to the caption. We note that upto a match threshold of 0.6, performance increases slightly, and there is no steep reduction in dataset size. After 0.7 however, the number of matches falls steeply as the match threshold is increased, leading to fewer videos and clips in the dataset, and a corresponding drop in downstream performance. We hence use a match threshold of $0.6$ to create our dataset.

\section{Method}\label{sec:model}
We focus on two different tasks in this paper that rely on video and text annotation - video retrieval and video captioning. We implement state of the art multimodal transformer models for each -- architectures and training objectives are defined in the next two sections. 
\subsection{Audiovisual Video Retrieval (AVR)} 
\begin{figure}[t]
\begin{center}
\includegraphics[width=\linewidth]{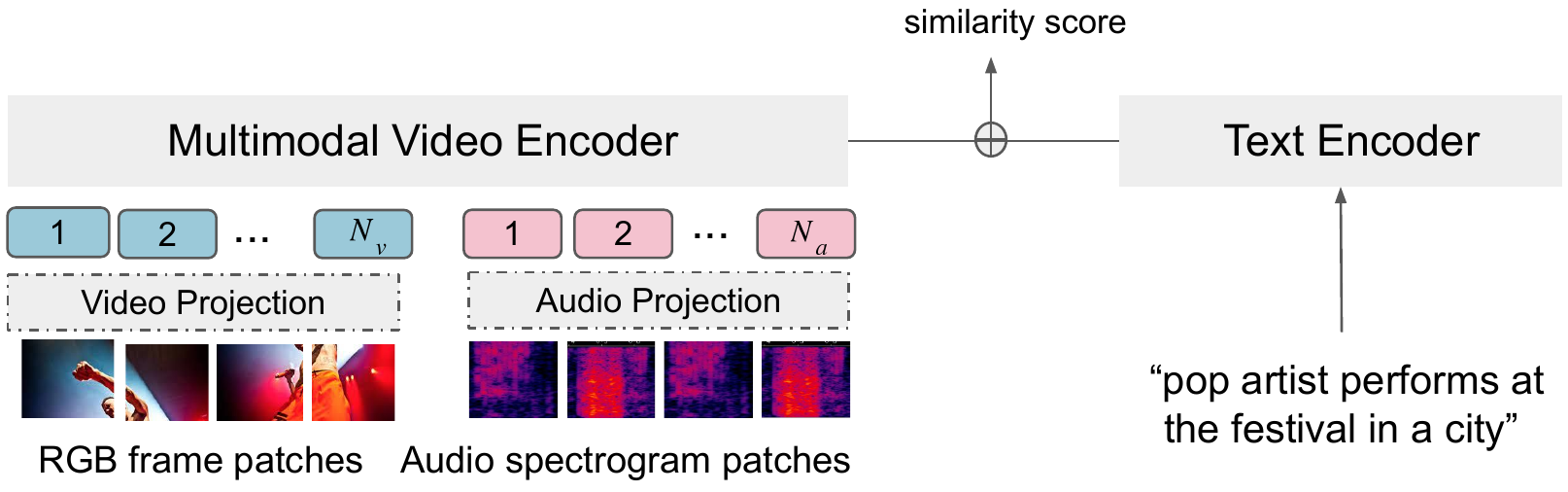}
\end{center}
\vspace{-1em}
\caption{\small{
\textbf{Our audiovisual dual stream retrieval model (AVR)}, which works for both image and audio focused retrieval datasets.
}}
    \vspace{-2em}
\label{fig:model}
\end{figure}
For retrieval, we use a dual-stream model (one stream being an audio-video encoder and one stream being a text encoder for the caption), which when trained with a contrastive loss allows for efficient text-video retrieval. Note that the efficient dual stream approach has also used by MIL-NCE~\cite{miech20endtoend} and FIT~\cite{bain2021frozen}, but unlike these works, our video encoder is multimodal (Fig. \ref{fig:model}), and utilises the audio as well. Our model is flexible, and can be used for audio-only, video-only and audio-visual retrieval.

\noindent\textbf{Multimodal Video Encoder:}
\label{sec:encoder}
Unlike recent works, we implement an audio-visual transformer based model that can be applied to both text-video and text-audio retrieval (figure in suppl material). Our encoder is inspired by the recently proposed MBT~\cite{nagrani2021attention}, which operates on RGB frames extracted at a fixed sampling rate from each video, and log-mel spectrograms used to represent audio. We first extract $N$ non-overlapping patches from the RGB image (or the audio spectrogram), similar to the way done by ViT~\cite{dosovitskiy2021an} and AST~\cite{gong2021ast} respectively.
The model consists of a number of transformer layers for each modality, with separate weights for each modality and fusion done via bottleneck tokens. Unlike MBT, we use frames extracted at a larger stride (an ablation is provided in the experiments), to cover the longer videos in retrieval datasets. We implement both RGB-only, audio-only and RGB-audio fusion models. \\
\noindent\textbf{Text encoder:} The text encoder architecture is the BERT model~\cite{devlin2019bert}. For the final text encoding, we use the [CLS] token output of the final layer. \\
\noindent\textbf{Joint embedding:}
For the final video encoding, we average the [CLS] tokens from both audio and RGB modalities. Both text and video encodings are then projected to a common dimension $D=256$ via a single linear layer each. We then compute the dot product similarity between the two projected embeddings after normalisation. \\
\noindent\textbf{Loss:}
We use the NCE loss~\cite{Zhai2019ClassificationIA} to learn a video and text embedding space, where matching text-video pairs in the batch are treated as positives, and all other pairwise combinations in the batch are treated as negatives. We minimise the sum of two losses,
video-to-text
and text-to-video~\cite{bain2021frozen}. At test time, inspired by FILIP~\cite{yao2021filip}, we sample $K$ clips equally spaced from the video, compare each one to the text embedding, and average the similarity scores. 
\subsection{Video Captioning} \label{sec:captioning_loss}
For video captioning, we use an encoder-decoder style generative model. Our video encoder is the same as the one used above for retrieval. \\ 
\noindent\textbf{Decoder:} To generate a text caption, we adapt the autoregressive GPT-2 (117M) decoder~\cite{radford2019language}, however we condition each predicted text token on video features from the video encoder as well as previously generated text tokens. More formally, given video features $C$ as context, to generate the next token $y_i$ in our caption $Y$, we first encode the previous generated tokens $Y_i=\{y_{0}, \dots, y_{i-1}\}$ with a look-up table and a positional embedding to produce $H_i=\{h_{0}, \dots, h_{i-1}\}$.
We then encode the context $C$ and the previous embedded tokens $H_i$ using a single transformer. The outputs of this transformer are $\Tilde{C} \cup \Tilde{H}_i$, where $\Tilde{H}_i=\{\Tilde{h}_0,\dots,\Tilde{h}_{i-1}\}$. We then predict the next token $y_i$ from $\Tilde{h}_{i-1}$ using a linear projection with a softmax: $ y_i = \mathrm{argmax}(\mathrm{softmax}(\Phi \Tilde{h}_{i-1}))$
where $\Phi \in \mathbb{R}^{\nu\times d}$ is the linear projection matrix and $\nu$ is the vocabulary size.
As is standard, the first word $h_0$ is set using a special \texttt{BOS} (beginning of sentence) token, and tokens are generated until a special \texttt{EOS} (end of sentence) token is generated. \\
\noindent\textbf{Loss:} We minimise the negative log-likelihood of generating the ground-truth caption~\cite{chen2019motion}.

\section{Experiments}
We evaluate our text-video models on the following tasks - text-video retrieval, which is video retrieval on primarily \textit{visual focused} datasets (Sec.~\ref{sec:retrieval}), text-audio retrieval, where captions are primarily focused on \textit{audio sounds},  (Sec.~\ref{sec:audio-retrieval}) and video captioning (Sec.~\ref{sec:captioning}). We use the common protocol of pretraining our models on a large dataset first, either VideoCC3M or HowTo100M, and then finetune on the target downstream dataset. Note that unlike other works, we apply the same same pretrained models for both visual-focused datasets such as MSR-VTT and audio-focused datasets such as AudioCaps and Clotho. We also investigate zero-shot performance, where we apply pretrained models directly to the target task, without any finetuning at all. In this case, no supervised video-text data is used at all.  We first describe datasets and metrics, then the implementation details, before finally discussing the results for each task.
\subsection{Datasets and Metrics} 
\noindent\textbf{VideoCC3M:} We use the VideoCC3M dataset created using our automatic mining method described in Sec. \ref{sec:mining-pipeline}. \\
\noindent\textbf{HowTo100M~\cite{miech2019howto100m}:} consists of 1.2M instructional videos. Weak captions are in the form of transcribed speech, which we obtain using the YouTube ASR API~\cite{youtubeapi}.\\
\noindent\textbf{MSR-VTT~\cite{xu2016msr}} contains
10K videos with 200K descriptions. For retrieval, we follow other works~\cite{Liu19a}, and train on 9K train+val videos, reporting results on the 1K-A test set. For captioning, we use the standard splits proposed in~\cite{xu2016msr}. \\
\noindent\textbf{AudioCaps~\cite{kim2019audiocaps}} is a dataset of video clips with
natural language captions that was introduced for the task of audio captioning, with clips sourced from the AudioSet dataset~\cite{gemmeke2017audio}. This dataset was then repurposed by~\cite{oncescu2021audio} for the task of text-audio retrieval, by taking a subset that does not overlap with the VGGSound~\cite{chen2020vggsound} dataset. After filtering out the videos no longer available on the web, we end up with  47,107 training, 403 validation and 778 test samples.\\
\noindent\textbf{Clotho~\cite{drossos2020clotho}} is an audio-only dataset of described sounds (with sounds sourced from the Freesound platform~\cite{font2013freesound}). During labelling, annotators only had access to the audio stream (no other meta tags or visual information). 
The data consists of a dev set and eval set of 2893 and 1045 audio samples respectively. Every audio sample
is accompanied by 5 captions. We follow ~\cite{oncescu2021audio} and treat each of the 5 captions per test audio as a separate query. 

\noindent\textbf{Metrics}
As is standard for retrieval, we report recall@K, $K \in \{1,5,10\}$. For captioning, we use the established metrics Bleu-4 (B-4)~\cite{papineni2002bleu}, CIDEr (C)~\cite{vedantam2015cider}, and Meteor (M)~\cite{banerjee2005meteor}. 
\subsection{Implementation details} 
In this section we describe implementation details for our models as well as certain design choices for sampling and initalisation. More details are provided in the appendix. \\
\noindent\textbf{Audio-visual encoder:} We use the ViT-Base (ViT-B, $L=12$, $N_H=12$, $d=3072$), as a backbone with $B=4$ fusion tokens and fusion layer $l_f=8$. We sample $32$ RGB frames for MSR-VTT, and $8$ RGB frames for AudioCaps. For audio we extract spectrograms of size $800 \times 128$ spanning $24$ seconds. \\
\noindent\textbf{Text encoder:} We use the BERT-Base architecture ($L=12$, $N_H=12$, $d=768$) with uncased wordpiece tokenization \cite{devlin2018bert}. We use a total number of $32$ tokens per caption during training -- cropping and padding for sentences longer and shorter respectively. No text augmentation is applied. \\
\noindent\textbf{Clip coverage:} A single segment per clip is randomly sampled at training time. We experiment with the length of this segment, controlled by the stride of the frames (32 frames at a stride of 2 frames at 25fps indicates an effective segment length of ~2.5 seconds).
We experiment with stride = {2, 6, 10, 14, 18}, and find optimal performance with stride = 14 frames (effective coverage of 18s). At test time, we sample $4$ clips equally spaced from the video, compare them to the text embedding, and average the similarity scores. More details are provided in the supplementary material. \\
\noindent\textbf{Video encoder initialisation:}
Unless otherwise specified, we use Kinetics-400~\cite{kay2017kinetics} initialisation for both video retrieval and captioning. For audio-focused retrieval datasets we initialise the model with VGGSound~\cite{chen2020vggsound} (see appendix).\\
\noindent\textbf{Training for retrieval:}
The temperature hyperparameter $\sigma$ for the NCE loss is set to 0.05, and the dimension of the common text-video projection space is set to 256. All models are trained with batch size $256$, synchronous SGD with momentum $0.9$, and a cosine learning rate schedule with warmup of $1.5$ epochs on TPU accelerators. For pretraining, we train models for $4$ epochs, and finetune for $5$ epochs. \\
\noindent\textbf{Training for captioning:}
We use the Adam optimizer with initial learning rate $1E-4$ and weight decay $0.01$. For all models, we pretrain for 120K iterations with a batch size of 512. For finetuning, we train for 1K iterations.

\begin{table}
\centering
\scalebox{0.85}{
\begin{tabular}{lcccc}
\toprule
\textbf{Init.} & \textbf{Modality} & \textbf{R@1} & \textbf{R@5} & \textbf{R@10} \\
\midrule
Scratch & V &   9.4 & 
22.5 & 
31.7 \\
ImageNet21K~\cite{deng2009imagenet} & V &  30.2 & 
59.7 & 
71.3 \\
K400~\cite{kay2017kinetics} & V &  30.2 & 
60.7 & 
71.1 \\
ImageNet21k~\cite{deng2009imagenet} & V+A &  32.2 & 
62.7 & 
74.4 \\
K400~\cite{kay2017kinetics} & V+A &  \textbf{32.3} & 
\textbf{64.1} & 
\textbf{74.6}\\
\bottomrule
\end{tabular}}
\caption{\textbf{Ablations for text-video retrieval on the MSR-VTT dataset.} Init. Initialisation of \textit{video encoder only}. Note we do not show audio-only results as some videos in the MSR-VTT dataset are missing audio. No VideoCC data is used here. Modalities are \textbf{V:} RGB, \textbf{A:} Audio spectrograms.}
\label{tab:msr-vtt-encoder}
\end{table}

\begin{table}
\centering
 \scalebox{0.87}{\begin{tabular}{lccccc}
\toprule
\textbf{Pretraining Data} & \textbf{Modality} &  \textbf{\# Caps} & \textbf{R@1} & \textbf{R@5} & \textbf{R@10} \\
\midrule
\textit{Finetuned} \\ 
- & V & - & 30.2 & 60.7 & 71.1\\
HowTo100M~\cite{miech2019howto100m} & V & 130M & 33.1
 & 62.3 & 
72.3 \\
VideoCC3M  & V & 970K & 35.0 & 
63.1 & 75.1 \\
VideoCC3M  & A+V & \textbf{970K} & \textbf{35.8} & 
\textbf{65.1} & \textbf{76.9} \\
\midrule 
\textit{Zero-shot} \\
HowTo100M~\cite{miech2019howto100m} & V & 130M & 8.6 & 16.9 & 25.8 \\
VideoCC3M & V &  970K & 18.9 & 
37.5 & 
47.1 \\
VideoCC3M & A+V &  \textbf{970K} & \textbf{19.4} & 
\textbf{39.5} & 
\textbf{50.3} \\
\bottomrule
\end{tabular}}
\caption{\textbf{Effect of pretraining data on text-video retrieval for the MSR-VTT dataset.} \textbf{\# Caps:} Number of unique captions. Training on VideoCC3M provides much better performance than Howto100M, with a fraction of the dataset size (VideoCC3M has only 970K captions and 6.3M clips compared to the 130M clips in HowTo100M) . The performance boost is particularly large for the zero-shot setting.}
\vspace{-1em}
\label{tab:msr-pretraining}
\end{table}
\begin{table}
\centering
\scalebox{0.82}{
\begin{tabular}{llcccc}
\toprule
\textbf{Method} &\textbf{Visual-Text PT} & \textbf{\# Caps} & \textbf{R@1} & \textbf{R@5} & \textbf{R@10} \\ \midrule
\textit{Finetuned} \\
HERO~\cite{li2020hero}  &HowTo100M & 136M & 16.8 & 43.4 & 57.7   \\
NoiseEst.~\cite{amrani2020noise}   &HowTo100M & 136M & 17.4 & 41.6 &  53.6   \\
 CE~\cite{Liu19a}$\dagger$ &   -  & & 20.9 & 48.8 & 62.4\\
UniVL~\cite{luo2020univilm} &HowTo100M & 136M & 21.2 & 49.6 & 63.1  \\ 
ClipBERT~\cite{lei2021less} &Coco, VisGen & 5.6M & 22.0 & 46.8 & 59.9   \\ 
AVLnet~\cite{rouditchenko2020avlnet} & HowTo100M & 136M & 27.1 & 55.6 & 66.6  \\
 MMT~\cite{gabeur2020multi}$\dagger$ &HowTo100M & 136M & 26.6 & 57.1 & 69.6 \\
 T2VLAD~\cite{wang2021t2vlad}$\dagger$ & - & & 29.5 & 59.0 & 70.1 \\
Support Set~\cite{patrick2020support}&HowTo100M & 136M & 30.1 & 58.5 &69.3   \\
VideoCLIP~\cite{xu2021videoclip} & HowTo100M & 136M & 30.9 & 55.4 & 66.8  \\
FIT~\cite{bain2021frozen} & CC3M & 3M & 25.5 & 54.5 & 66.1\\
FIT~\cite{bain2021frozen} & Multiple$\ddagger$ & 6.1M & 32.5 & 61.5 & 71.2\\
\textbf{Ours}  & VideoCC3M & \textbf{970K} & \textbf{35.8} & 
\textbf{65.1} & 
\textbf{76.9}  \\
\midrule
\textit{Zero-shot} \\
MIL-NCE~\cite{miech2019howto100m} &HowTo100M & 136M & 7.5 & 21.2 & 29.6    \\
SupportSet~\cite{patrick2020support} & HowTo100M & 136M & 8.7 & 23.0 & 31.1     \\
EAO~\cite{shvetsova2021everything} & HT100M & 136M & 9.9 & 24.0 & 32.6 \\
VideoCLIP~\cite{xu2021videoclip} & HowTo100M & 136M & 10.4 & 22.2 &  30.0 \\
FIT~\cite{bain2021frozen} & WebVid2M* & 2.5M & 15.4 & 33.6 &  44.1  \\
\textbf{Ours} & VideoCC3M & \textbf{970K} &  \textbf{19.4} & 
\textbf{39.5 }& 
\textbf{50.3} \\
\bottomrule

\end{tabular}}
\caption{\label{tab:msr-vtt-sota} \textbf{Comparison to state-of-the-art results on MSR-VTT 1k-A split for text-to-video retrieval.} \textbf{Visual-Text PT:} Visual-text pretraining data. \textbf{\# Caps:} Number of unique captions used during pretraining. $\dagger$ These works use numerous experts, including Object, Motion, Face, Scene, Speech, OCR and Sound classification features. $\ddagger$ Pretrained on WebVid-2M, CC3M and COCO datasets. *Numbers obtained from the authors.}
\vspace{-1em}
\end{table}

\subsection{Text-video Retrieval} \label{sec:retrieval}
\noindent\textbf{Video encoder initialisation:} We first experiment with initalising the video encoder \textit{only} (Table \ref{tab:msr-vtt-encoder}, and find that while ImageNet initalisation provides a significant boost over training from scratch, using Kinetics-400 (K400) for video only provides a very marginal further gain. This suggests that at least for retrieval, the initialisation of the video encoder is not as important as joint text-video pretraining for the entire model (as demonstrated in the next paragraph).  \\
\noindent\textbf{Effect of pretraining data:}
We begin by analysing the results with fine-tuning for text-video retrieval on the MSR-VTT dataset, presented in Table \ref{tab:msr-pretraining}. We note that pretraining on VideoCC3M provides a significant boost to performance over HowTo100M, with far less data, and for an RGB-only model, yields a 5\% improvement over training from scratch on R@1.
This effect is even more profound in the  zero-shot case, where for an RGB-only model, using VideoCC3M more than doubles the R@1 performance compared to HowTo100M pretraining. This is done with 100x fewer captions and 20x less video data. We believe that this shows the value in high-quality video-captioning pairs.
Regarding audio inputs, we note that MSR-VTT is a visual benchmark (unlike AudioCaps and Clotho), with some videos missing an audio track entirely. However we show that adding audio provides a modest performance boost. 
We then compare to previous works on this dataset in Table \ref{tab:msr-vtt-sota}, including recently released Frozen In Time (FIT)~\cite{bain2021frozen} and VideoCLIP~\cite{xu2021videoclip}.
We note that our model outperforms FIT which pretrains on 3 different datasets - CC3M, WebVid2M and COCO~\cite{chen2015microsoft}. We were unable to train on WebVid2M due to data restrictions but believe further performance gains could be achieved by training on VideoCC3M and WebVid jointly. We also note that by training on VideoCC3M, we outperform FIT trained only on the CC3M dataset by a big margin (R@1 25.5 to 35.3), even though  the amount of manually annotated supervision is the same. This shows the benefit of mining extra video data using our data mining pipeline. On zero-shot performance, we outperform all previous works that pretrain on HowTo100M, and FIT~\cite{bain2021frozen} when it is trained only on video data (WebVid2M). We note that adding in various image datasets provides a huge boost to performance in FIT~\cite{bain2021frozen}, and this complementary approach could be used with our dataset. We could also use additional seed datasets such as COCO Captions~\cite{chen2015microsoft} to mine more text-video clips, which we leave as future work. \\
\noindent\textbf{Results using CLIP~\cite{radford2021learning}}
Given the recent flurry of CLIP based~\cite{luo2021clip4clip,gao2021clip2tv,cheng2021improving,fang2021clip2video}, RGB-only works for video retrieval, in this section we show the complementarity of using CLIP~\cite{radford2021learning} based models trained on the 400M pair WiT dataset such as Clip4Clip~\cite{luo2021clip4clip} finetuned on the VideoCC dataset. 
We reproduce Clip4Clip~\cite{luo2021clip4clip} with mean pooling in our framework (Table \ref{tab:clip4clip}). Using CLIP (trained on 400M diverse image-caption pairs) leads to very strong zero-shot performance, however finetuning it on VideoCC \textit{further} improves performance by over 3\% R$@$1, showing the additional value of automatically mined \textit{videos}. We also outperform the zero-shot SOTA from Clip4Clip which was post trained on a curated subset of HowTo100M and is the highest online number for this zero-shot benchmark (CaMoE~\cite{cheng2021improving} and Clip2TV~\cite{gao2021clip2tv} do not report zero-shot results). This shows the value of our automatic video mining pipeline.
\vspace{-1em}
\begin{table}[h]
    \centering
    \begin{tabular}{lcccc}
    \toprule
     Model  & PreTraining Data & $R@1$ & $R@5$ & $R@10$ \\
     \midrule
    C4C~\cite{luo2021clip4clip} & WiT~\cite{radford2021learning} &30.6 & 54.4 & 64.3 \\ 
    Ours & WiT~\cite{radford2021learning} + VideoCC & \textbf{33.7} & \textbf{57.9} & \textbf{ 67.9}\\
     \bottomrule
    \end{tabular}
    \caption{Finetuning Clip4Clip~\cite{luo2021clip4clip} (C4C) on VideoCC for zero-shot performance on MSR-VTT.
    }
    \label{tab:clip4clip}
\end{table}

\begin{table}
\centering
\scalebox{0.87}{
\begin{tabular}{lcccc}
\toprule
\textbf{Model} & \textbf{Pretraining} & \textbf{Modality} & \textbf{R@1} & \textbf{R@10} \\
\midrule
SOTA~\cite{oncescu2021audio}$\dagger$ & - & A &24.3
& 72.1\\ 
Ours & - & A & 32.0 & 82.3 \\
Ours & HowTo100M & A & 33.7 & 83.2\\
Ours & VideoCC3M & A & 35.5 & 84.5\\
\rowcolor{aliceblue} Ours (ZS) & HowTo100M & A & 1.4 & 6.5 \\
\rowcolor{aliceblue} Ours (ZS) & VideoCC3M & A & 8.7 & 37.7 \\
\midrule
SOTA~\cite{oncescu2021audio}$\dagger$ & - &  A+V &28.1
& 79.0\\ 
Ours & - &  A+V & 
41.4 &
85.3   \\
Ours & VideoCC3M & A+V & \textbf{43.2} & \textbf{88.9}\\
\rowcolor{aliceblue} Ours (ZS) & VideoCC3M & A+V & 10.6 & 45.2 \\ 
\bottomrule
\end{tabular}}
\caption{\textbf{Results on the AudioCaps dataset for text-audio retrieval.} $\dagger$ Higher than reported in the paper, as these are provided by authors on our test set. Inputs refers to video inputs as follows: \textbf{A:} Audio spectrograms \textbf{V:} RGB video frames. \protect\MA{Rows highlighted in light blue show Zero-shot (ZS) performance.} }
\label{tab:audiocaps}
\end{table}
\begin{table}
\centering
\scalebox{0.87}{
\begin{tabular}{lccc}
\toprule
\textbf{Model} & \textbf{Pretraining} & \textbf{R@1} & \textbf{R@10} \\
\midrule
SOTA~\cite{oncescu2021audio} & - & 6.7 & 33.3 \\
Ours & - & 7.8 & 35.4 \\ 
Ours & VideoCC3M & \textbf{8.4} & \textbf{38.6}\\
\rowcolor{aliceblue} Ours (ZS) & VideoCC3M & 3.0 & 17.5 \\
\midrule
SOTA~\cite{oncescu2021audio} & AudioCaps  & 9.6 & 40.1\\ 
Ours & AudioCaps & 11.4 & 43.4\\
Ours & VideoCC3M+AudioCaps & \textbf{12.6} & \textbf{45.4}\\
\bottomrule
\end{tabular}}
\caption{\textbf{Results on the Clotho dataset for text-audio retrieval.} \protect\MA{Rows highlighted in light blue show Zero-shot (ZS) performance.} Note this dataset contains audio only (no RGB frames).}
\label{tab:clotho}
\end{table}

\subsection{Audio Retrieval}\label{sec:audio-retrieval}
\begin{table}[t]
    \centering
    \scalebox{0.8}{
    \begin{tabular}{@{\hskip 0.8mm}ll@{\hskip 2.5mm}c@{\hskip 3mm}c@{\hskip 3mm}c@{\hskip 3mm}c@{\hskip 3mm}c@{\hskip 0.8mm}}
    \toprule
        \textbf{Method}  & \textbf{PT} & \textbf{Modality} & \textbf{B-4} & \textbf{C} & \textbf{M} \\
        \midrule 
        \textit{Finetuned} \\ 
        POS+CG~\cite{wang2019controllable} &-& V & 42.00 & 49 & 28.20 \\
        POS+VCT~\cite{hou2019joint} &-& V & 42.30 & 49 & 29.70 \\
        SAM-SS~\cite{chen2020semantics} &-& V & 43.80 & 51 & 28.90\\
        ORG-TRL~\cite{zhang2020object} &-& V & 43.60 & 51 & 28.80 \\
        VNS-GRU~\cite{chen2020delving} &-& V & 45.30 & 53 & 29.90 \\ 
        UniVL~\cite{luo2020univl}  &HowTo100M& V+T & 41.79 & 50 & 28.94  \\
        
        DECEMBERT~\cite{tang2021decembert} &HowTo100M& V & 45.20 & 52 & 29.70  \\ 
        Ours &HowTo100M& V & \bf 47.33  & 55 & \textbf{37.11} \\ 
        Ours &VideoCC3M& V & 45.47  & \textbf{55} & 36.96  \\ 
        \midrule 
     \textit{Zero-shot} \\    
Ours & HowTo100M & V & 7.5 &
0.5 & 
8.23 \\
Ours & VideoCC3M & V & \textbf{13.23} & 
\textbf{8.24} & 
\textbf{11.34}  \\
\bottomrule
    \end{tabular}}
    \caption{\textbf{Results on the MSR-VTT dataset for video captioning.} Zero-shot results are obtained without any annotated video-text data. Modalities: \textbf{V:} RGB frames. \textbf{T:} ASR in videos.}
    \vspace{-1em}
    \label{tab:sota-msr-captioning}
\end{table}
\begin{figure*}[th]
\begin{center}
\includegraphics[width=0.93\linewidth]{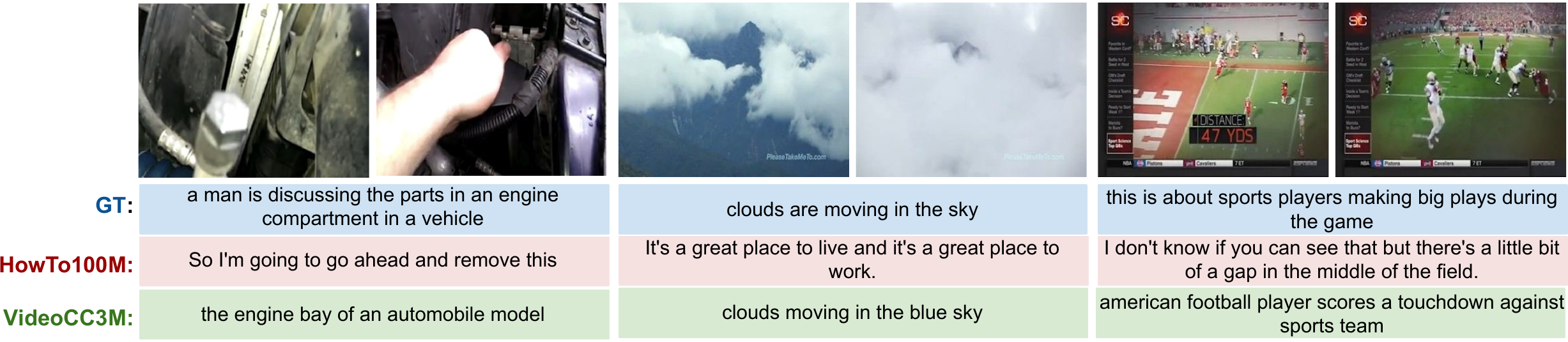}
\end{center}
\vspace{-1em}
\caption{
\textbf{Zero-shot captioning results on MSR-VTT test set videos.} We show 2 frames per clip. As expected, the style of the predicted captions from a model pretrained on HowTo100M are similar to ASR, and often concepts are tenuously related (middle example). Pretraining on VideoCC3M yields captions that are closer to the ground truth.
}
\label{fig:zero-shot-captioning}
\end{figure*}
For text-audio retrieval we report results on two
audio-centric datasets (i.e.\ datasets paired with natural language descriptions that focus explicitly
on the content of the audio track) - AudioCaps~\cite{kim2019audiocaps} and Clotho~\cite{drossos2020clotho}. The goal here is to retrieve the correct audio segment given a free form natural language query. While Clotho comes with only audio, AudioCaps has both audio and RGB frames. 

Results on the AudioCaps dataset are provided in Table \ref{tab:audiocaps}. We first show results for an audio-only encoder (we only feed spectrograms as input). We note that our model with no audio-text pretraining already outperforms the current state of the art~\cite{oncescu2021audio} by a large margin (R@1: from 24.3 to 32.0), despite the fact that~\cite{oncescu2021audio} uses features pretrained on VGGSound and VGG-ish features pretrained on YouTube8M. This could be because unlike their encoder, our encoder is trained end-to-end directly from spectrograms. We then show results with pretraining on the spectrograms from HowTo100M (no RGB frames are used here), and find that there is some improvement. Pretraining on the audio and captions from VideoCC3M however, gives substantial performance gains to R@1 by over 3\%. This improvement is particularly impressive because the captions were transferred via visual similarity to still images and no additional manual audio-text supervision was used.

We also report zero-shot results, and find that unsurprisingly, pretraining on HowTo100M results in poor performance, likely because the model has learned to focus on speech. VideoCC3M provides a large improvement, however there is still a distance to finetuning performance.

Finally, we also show that using an audio-visual fusion encoder and training on VideoCC3M provides a further significant improvement  demonstrating the complementarity of RGB information for this task. 

Results on Clotho are provided in Table \ref{tab:clotho}. Here we show a similar trend, however this dataset is more challenging. Because Clotho is also a much smaller dataset, we also show results with AudioCaps pre-training as is done by~\cite{oncescu2021audio}. Combining AudioCaps supervised pretraining after VideoCC3M pretraining provides the best result.

\subsection{Video Captioning}\label{sec:captioning}
Results for video captioning are provided in Table \ref{tab:sota-msr-captioning}. For finetuning, we note that our model pretrained on VideoCC3M outperforms previously published works. Unlike retrieval, we note that finetuning on the HowTo100M dataset provides slight gains to the B-4 and M metrics, but VideoCC3M is still competitive with a fraction of the data size. 
We then compare zero-shot performance, and find that pretraining on HowTo100M performs poorly, potentially because of the large difference in style and domain between instructional speech and human-generated captions. Training on VideoCC3M provides a substantial boost across all metrics, again with a fraction of the training data. Some qualitative results are shown in Fig. \ref{fig:zero-shot-captioning}.

\section{Conclusion}

We propose a new, automatic method for leveraging existing image datasets to mine video and audio data with captions. We apply it to the CC3M dataset~\cite{sharma2018conceptual} to mine millions of weakly labelled video-text pairs. Our mining pipeline is scalable, and will be applied to even larger image datasets such as YFC100M~\cite{thomee2016yfcc100m}. Training a multimodal retrieval model on these clips leads to state of the art performance for video retrieval and captioning, and shows complementarity with existing image-text models such as CLIP. Future work can focus on augmenting these automatic captions with even more video related text, such as action labels.\\
\noindent\textbf{Societal Impact:} We note that transformers are in general compute-heavy, which can have adverse environmental effects. We believe that releasing a dataset that is an order of magnitude smaller than HowTo100M, but provides better zero-shot generalisation, will lead to faster and cheaper language-video model innovation. Finally, our dataset may reflect biases present in videos online, as well as biases in the captions of the seed dataset. Existing biases may render models trained on this data unsuitable for certain applications. It is important to keep this in mind when deploying, analysing and building upon these models. \\
\noindent\textbf{Fairness Analysis on the Data:} We start with input data, CC3M~\cite{sharma2018conceptual}, that has already tried to mitigate fairness issues. This data source has many fewer fairness issues than than website scraping efforts focusing on scale as evaluated in~\cite{birhane2021multimodal}. We made further efforts to mitigate fairness issues in the text domain, image domain, and video domain, by performing both automated and manual analysis.
For automated analyses, we evaluated the text using NLP tools for toxicity and PII, while images and videos were reviewed for their likelihood of containing mature or offensive imagery.
For manual analyses, we inspected thousands of caption-video pairs where the captions contained words that are sensitive or have been previously shown to have fairness disparities such as those listed in~\cite{birhane2021multimodal,yang2020towards}, and provided at least some further mitigation of extreme errors.

{
    \clearpage
    \small
    \bibliographystyle{ieee_fullname}
    \bibliography{macros,main}
}
\clearpage

\appendix
\addcontentsline{toc}{section}{Appendix} 
\part{Appendix} 
{\hypersetup{linkcolor=black} \parttoc}

\section{VideoCC3M dataset}
In this section we provide some more details on the automatically mined clips that are part of the VideoCC3M dataset, including basic statistics, more qualitative examples, and a brief human study to assess the quality of the mined clips.
\subsection{Dataset statistics}
\begin{table*}[h!]
\centering
\caption{\textbf{Dataset Statistics:} VideoCC3M is an order of magnitude larger than existing video-text datasets in the number of videos and captions. Rows highlighted in blue are large-scale, weakly annotated datasets. WVT uses titles and descriptions from YouTube videos, and HowTo100M has noisy text supervision from ASR. $\dagger$ Not publicly released.}
\footnotesize
\begin{tabular}{llrrrrr}
\toprule
dataset & domain & \# clips & average clip length (s) & \# captions & time (hr) & \# pairs \\ \midrule
MPII Cook~\cite{rohrbach2012database} & cooking & 44 & 600 & 6K & 8 & 6K\\ 

TACos~\cite{regneri2013grounding} & cooking & 7K & 360 & 18K & 15.9 & 18K \\ 

DideMo~\cite{anne2017localizing} & flickr & 27K & 28 & 41K & 87 & 41K\\ 
MSR-VTT~\cite{xu2016msr} & open & 10K & 15 & 200K & 40 & 200K \\ 
Charades~\cite{sigurdsson2016hollywood} & home & 10K & 30 & 16K & 82 & 16K \\ 
LSMDC15~\cite{rohrbach2017movie} & movies & 118K & 4.8 & 118K & 158 & 118K \\ 
YouCook II~\cite{zhou2018towards} & cooking & 14K & 316 & 14K & 176 & 14K\\ 
ActivityNet~\cite{krishna2017dense} & action focused & 100K & 180 & 100K & 849 & 100K\\ 
CMD~\cite{bain2020condensed} & movies & 34K & 132 & 34K & 1.3K & 34K \\
WebVid-2M & open  & 2.5M & 18 & 2.5M & 13K & 2.5M \\ 
\midrule
\textbf{VideoCC3M }& \textbf{open} & \textbf{6,323,992} & \textbf{10} & \textbf{974,247}  & \textbf{17.5K} & \textbf{10,339,249} \\
\midrule 
\rowcolor{aliceblue} WVT~\cite{stroud2020learning}$\dagger$ & action focused & 70M & 10 & 70M & 194K & 70M\\
\rowcolor{aliceblue}
HowTo100M~\cite{miech2019howto100m} & instruction & 136M & 4 & 136M &  134.5K & 136M \\

\bottomrule            
\end{tabular}
\label{tab:datastats-stats}
\end{table*}

\begin{figure}[h]
    \centering
    \includegraphics[width=0.45\textwidth]{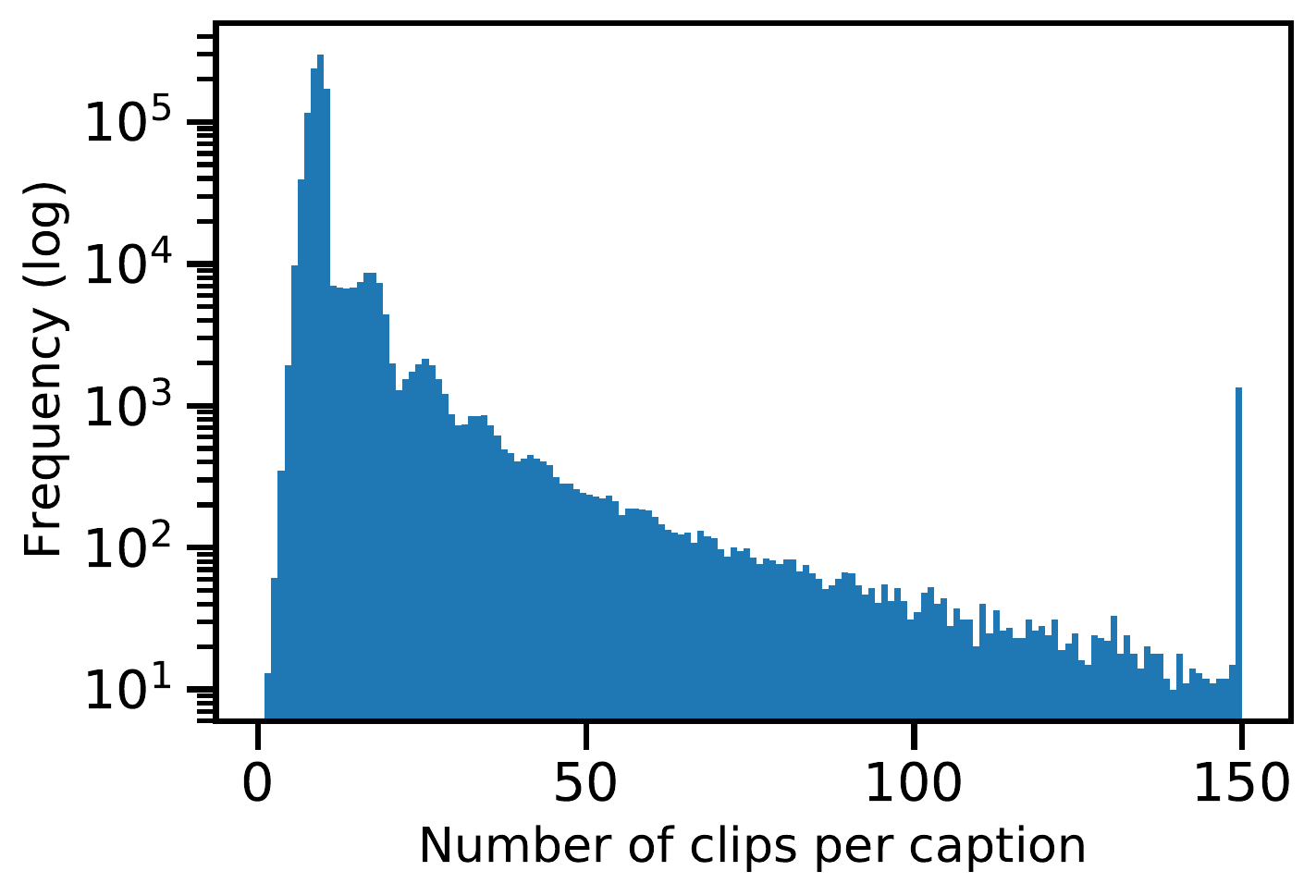}
    \caption{\textbf{Distribution of clips per caption in VideoCC3M.} Frequency of samples (y-axis) is on a log-scale. Because the CC3M dataset has some seed images that share the same caption, one caption can have more than 10 mined clips. 96.6K out of 97K captions have less than 50 clips per caption. All samples with more than 150 clips per caption are grouped into a single bin. }
    \label{fig:captions-hist}
\end{figure}

We provide the total number of unique captions, video clips and pairs in Table \ref{tab:datastats-stats} comparing VideoCC3M to other existing video and text datasets. Note that at 10M pairs, our dataset is much larger than manually annotated datasets but still much smaller than the large HowTo100M dataset. The full distribution of clips per caption is provided in Fig. \ref{fig:captions-hist}, (note that the y-axis is on a log scale). Each caption is matched to a mean of 10.6 clips, with some captions matched to more than 10 clips. This is possible because, while we limit the clip mining to 10 clips per seed image, the original CC3M dataset has multiple seed images with the same caption, eg `an image of digital art', leading to more than 10 mined clips for these captions.  96.6K out of 97K captions have less than 50 clips per caption. This added redundancy is an interesting feature of the data where visually similar clips share the same caption from the same seed image and visually distinct clips share that same caption from different seed images.

\subsection{Domains}
\begin{figure*}[t]
    \centering
    \includegraphics[width=1\textwidth]{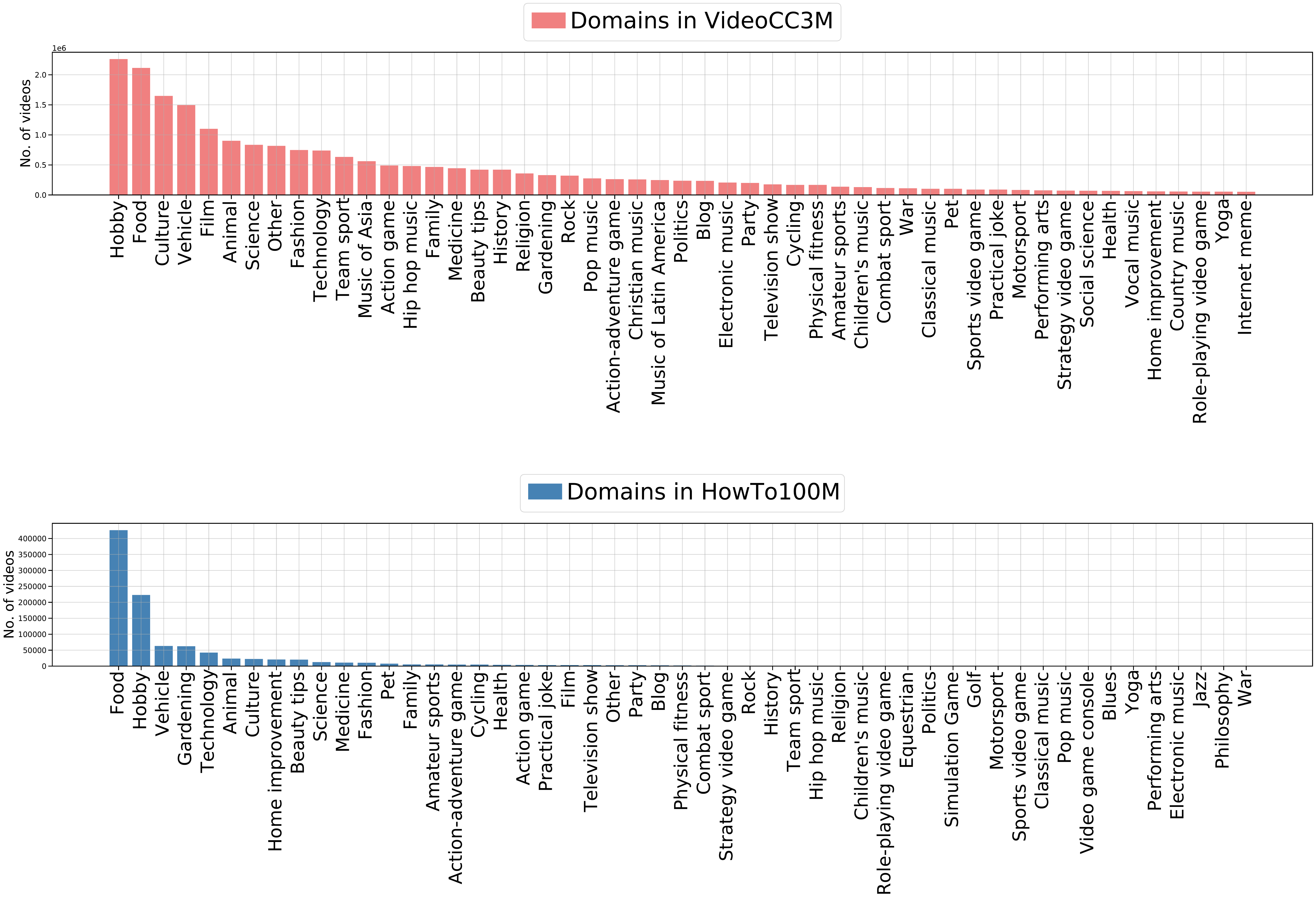}
    \caption{\textbf{Domains in VideoCC3M vs HowTo100M.} We show the top 50 domains for each dataset and group remaining samples into `Other'. Note how the domains in VideoCC are more balanced. Note HowTo100M has about 1M videos in the dataset.}
    \label{fig:domains-full}
\end{figure*}

We show the top 50 domains in Fig.~\ref{fig:domains-full} for both the VideoCC3M and the HowTo100M datasets, and group remaining samples into the `Other' domain. This figure expands the analysis presented in Figure~\ref{fig:domains} of the main paper. It is clear that the domains in VideoCC3M are more balanced, while HowTo100M videos are largely dominated by the `Food' and `Hobby' domains. This is unsurprising given that HowTo100M is limited to instructional videos.

\subsection{Human study on quality}
In order to quantiatively assess the quality of the mined clips in VideoCC3M, we also perform a quick manual assessment of 100 randomly sampled clips from the dataset. For each clip, we first annotate whether there is at least one frame in the clip matching the caption, and find that 91 out of 100 clips were labelled to have this property. We noticed that clips without a single frame matching the caption are often those where the seed image does not match the caption either, due to noise in the CC3M dataset. We then devise a simple quality score with the following scale of 3 values: 0 - not relevant, 1 - somewhat relevant, 2 -  very relevant, to assess the degree to which the caption matches the retrieved sample. For examples of clips that are somewhat relevant, see Fig. \ref{fig:mining-examples-negatives}. Over 100 samples, we get an average score of 1.51, with 9 samples having score 0, 31 having score 1 and 60 having score 2.

\subsection{More qualitative examples} 
\begin{figure*}[t]
\begin{center}
\includegraphics[width=0.95\linewidth]{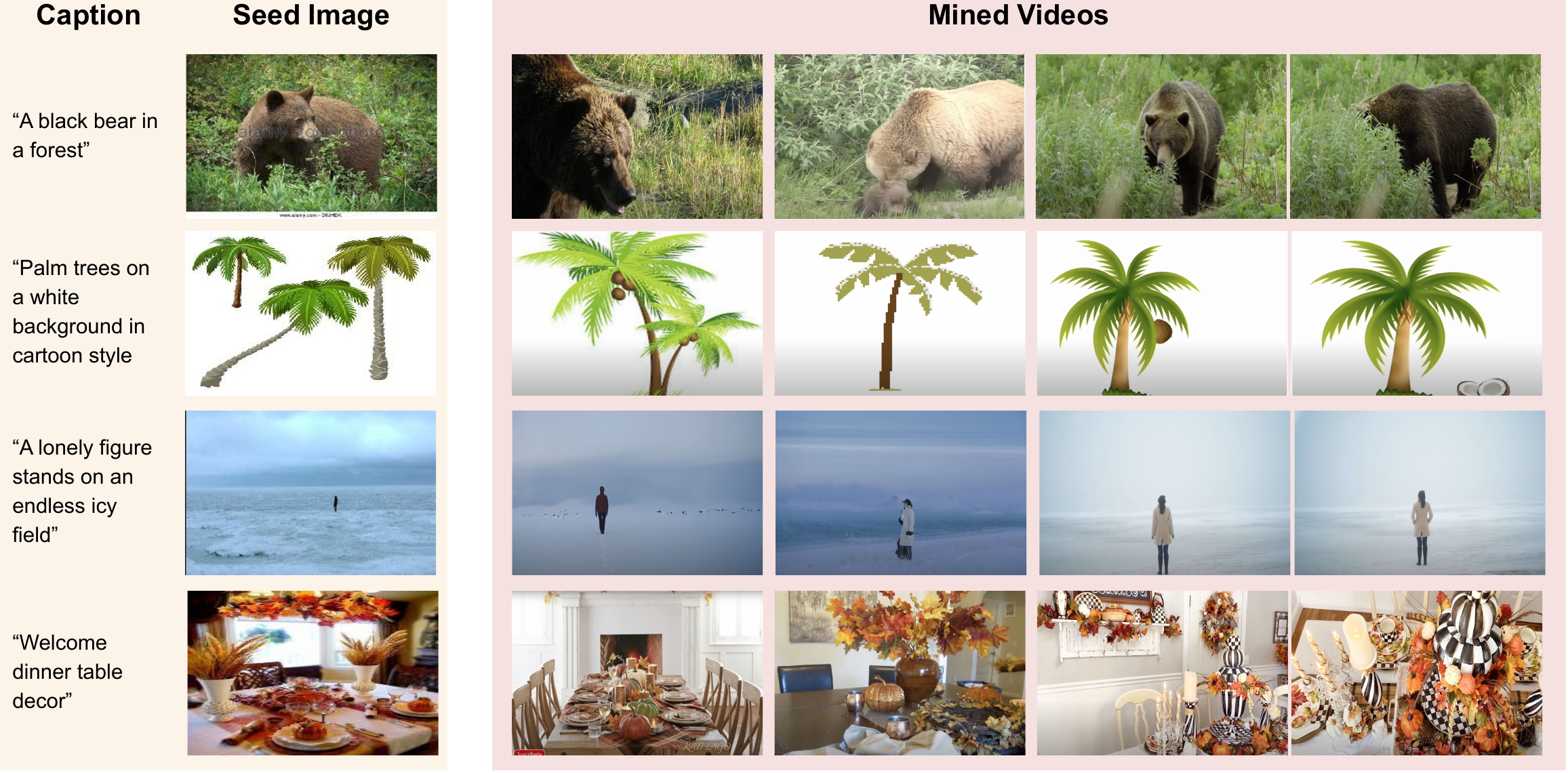}
\end{center}
\caption{
\textbf{Examples of clips with captions that are mined automatically.} For each seed image, we show 3 `matched' clips obtained using our automatic video mining method. For the first 2 clips, we show only a single frame, but for the third clip we present 2 frames to show motion , either of the subjects in the video (first 3 rows - the bear, the coconut falling, the arms of the woman) or camera motion (last row). Note frames may have been cropped and resized for ease of visualisation.
}
\label{fig:mining-examples-more}
\end{figure*}

\begin{figure*}[t]
\begin{center}
\includegraphics[width=0.95\linewidth]{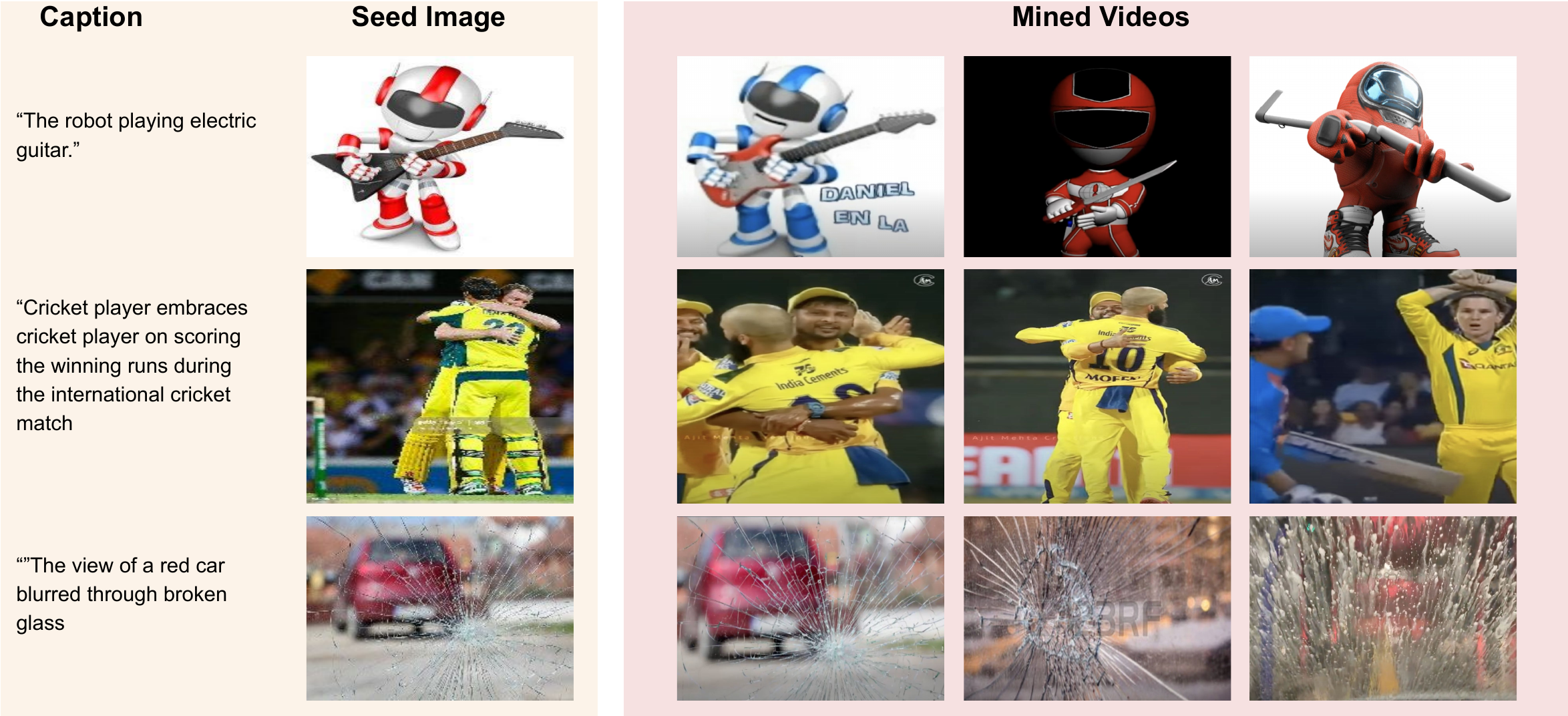}
\end{center}
\caption{
\textbf{Failure Cases: examples of somewhat related clips with captions that are mined automatically.} For each seed image, we show 3 `matched' clips obtained using our automatic video mining method. Here we show failure cases, where the matched clips are somewhat relevant to the caption, but not entirely. For example, top row - in the last two clips the robot are holding a long object but it is not a guitar, second row - last clip contains cricketers but they are not hugging. Finally in the third row, note that the second clip has the broken glass but no red car, whereas the last clip has a red car but the glass is not broken, it is being washed in a car wash. Note that the original seed image of the red car is originally from a video, which we retrieve using our pipeline. Note frames may have been cropped and resized for ease of visualisation.
}
\label{fig:mining-examples-negatives}
\end{figure*}

We show some more qualitative examples in Fig. \ref{fig:mining-examples-more}. Note the diversity of retrieved samples, including an animated video of a tree on a white background. In Fig. \ref{fig:mining-examples-negatives}, we also show some failure cases, where the clips are somewhat related to the captions but not perfectly. 

\subsection{Ablation on temporal length $t$}
We show the effect of the length of the mined clips on zero-shot performance on the MSR-VTT dataset. Results are in Table \ref{tab:temporal-span}. Although we know that video content diverges the further we are from the matched frame to the seed image, we find increasing the span actually increases performance up until 10 seconds. This is perhaps because videos tend to be correlated over time.  Unrelated extra information could also act as a regularisation, wherein slight noise does not harm the results. We hence use clips of 10 seconds in all further experiments with VideoCC3M, but we note that future work will more intelligently determine the boundary of the mined clips. 
\begin{table}
\centering
\begin{tabular}{lccccc}
\toprule
$t (s)$ & 3 & 5 & 10 & 20 & 30 \\ \midrule
MSR-VTT (ZS) & 16.4 & 17.1 & 18.9 & 18.8 & 18.8  \\
\bottomrule
\end{tabular}
\caption{
\textbf{Temporal Span $t$ of the mined clips.} We report zero-shot R@1 performance on the MSR-VTT dataset. 
} 
\label{tab:temporal-span}
\end{table} 
\section{VideoCC12M dataset}
We ran our mining pipeline with an additional seed image captioning dataset called Conceptual Captions 12M~\cite{changpinyo2021conceptual} (CC12M). CC12M is the recently released extension of Conceptual Captions 3M~\cite{sharma2018conceptual} (CC3M), . Note that while CC3M consists of higher quality captions~\cite{sharma2018conceptual}, CC12M was created by relaxing the data collection pipeline used in CC3M, and hence the captions are far noisier. VideoCC3M consists of 10.3M clip-text pairs from 6.3M video clips and 970K unique captions, while VideoCC12M contains 48.0M clip-text pairs from 30.3M video clips and 5.7M unique captions. While we include results on VideoCC12M for completeness, we note that for most tasks VideoCC3M is sufficient for good performance with far less data.

\subsection{Video Retrieval using VideoCC12M}
We show results in Table \ref{tab:msr-pretraining-with-12M}. Pretraining on the VideoCC12M dataset provides a further boost to performance over the VideoCC3M pretraining, particularly for R@10 and R@5. This furthers the improvement over the state of the art, which was provided in Table 3 in the main paper. Our model trained on VideoCC12M achieves R@1 37.1 compared to FIT~\cite{bain2021frozen}, which gets an R@1 of 32.5. Note FIT is pretrained on WebVid2M, COCO and CC3M. 

\begin{table}
\centering
 \scalebox{0.87}{\begin{tabular}{lccccc}
\toprule
\textbf{Pretraining Data} & \textbf{Modality} &  \textbf{\# Caps} & \textbf{R@1} & \textbf{R@5} & \textbf{R@10} \\
\midrule
\textit{Finetuned} \\ 
- & V & - & 31.2 & 60.7 & 71.1\\
HowTo100M~\cite{miech2019howto100m} & V & 130M & 33.1
 & 62.3 & 
72.3 \\
VideoCC3M  & V & 970K & 35.0 & 
63.1 & 75.1 \\
VideoCC3M  & A+V & 970K & 35.3 & 
65.1 & 76.9 \\
VideoCC12M & V & 5.7M & 36.9 & 
66.5 & 
75.6  \\
VideoCC12M & A+V & 5.7M & \textbf{37.1} & 
\textbf{67.5} & 
\textbf{77.6}  \\
\midrule 
\textit{Zero-shot} \\
HowTo100M~\cite{miech2019howto100m} & V & 130M & 8.6 & 16.9 & 25.8 \\
VideoCC3M & V &  970K & 18.9 & 
37.5 & 
47.1 \\
VideoCC3M & A+V &  970K & 19.4 & 
39.5 & 
50.3 \\
VideoCC12M & V & 5.7M &21.8 & 
44.5 & 
54.1 \\
VideoCC12M & A+V & 5.7M & \textbf{22.3} & 
\textbf{45.8} & 
\textbf{57.2} \\
\bottomrule
\end{tabular}}
\caption{\textbf{Effect of pretraining data on text-video retrieval for the MSR-VTT dataset.} \textbf{\# Caps:} Number of unique captions. Training on VideoCC provides much better performance than Howto100M, at a fraction of the dataset size, particularly for the zero-shot setting.}
\label{tab:msr-pretraining-with-12M}
\end{table}

\subsection{Video Captioning using VideoCC12M}
Results for video captioning are provided in Table \ref{tab:sota-msr-captioning-12M}. With the additional data from VideoCC12M, we are on par with HowTo100M in the finetuning setting. For the zero-shot setting, training on VideoCC3M provides a substantial boost across all metrics, with a fraction of the training data, and also outperforms training on VideoCC12M in the zero-shot setting. This is interesting, and we hypothesise it is because the captions in CC3M are far cleaner than CC12M. We note the exact same trend was reported for zero-shot image captioning in the CC12M paper~\cite{changpinyo2021conceptual}. This suggests that zero-shot performance depends more on the transferred caption quality, and future work may improve transfer performance by cleaning up captions in larger data sets. This finding reinforces the theme that more data is not always better.
\begin{table}[t]
    \centering
    \scalebox{0.8}{
    \begin{tabular}{@{\hskip 0.8mm}ll@{\hskip 2.5mm}c@{\hskip 3mm}c@{\hskip 3mm}c@{\hskip 3mm}c@{\hskip 3mm}c@{\hskip 0.8mm}}
    \toprule
        \textbf{Method}  & \textbf{PT} & \textbf{Modality} & \textbf{B-4} & \textbf{C} & \textbf{M} \\
        \midrule 
        \textit{Finetuned} \\ 
        ORG-TRL~\cite{zhang2020object} &-& V & 43.60 & 51 & 28.80 \\
        VNS-GRU~\cite{chen2020delving} &-& V & 45.30 & 53 & 29.90 \\ 
        UniVL~\cite{luo2020univl}  &HowTo100M& V+T & 41.79 & 50 & 28.94  \\
        
        DECEMBERT~\cite{tang2021decembert} &HowTo100M& V & 45.20 & 52 & 29.70  \\ 
        Ours &VideoCC3M& V & 45.47  & 55 & 36.96  \\ 
        Ours &HowTo100M& V & \bf 47.33  & 55 & 37.11 \\ 

        \bf Ours &VideoCC12M& V & 47.21  & \bf 56 & \bf 37.70 \\ 
        \midrule 
     \textit{Zero-shot} \\    
Ours & HowTo100M & V & 7.5 &
0.5 & 
8.23 \\
Ours & VideoCC3M & V & \textbf{13.23} & 
\textbf{8.24} & 
\textbf{11.34}  \\
Ours & VideoCC12M & V & 10.09 & 
3.58 & 
9.68 \\
\bottomrule
    \end{tabular}}
    \caption{\textbf{Results on the MSR-VTT dataset for video captioning.} Zero-shot results are obtained without any annotated video-text data. Modalities: \textbf{V:} RGB frames. \textbf{T:} ASR in videos.}
    \label{tab:sota-msr-captioning-12M}
\end{table}
\begin{table}
\centering
\begin{tabular}{lcccc}
\toprule
Stride & Span (s) & R@1 & R@5 & R@10 \\
\midrule
2 & 2.56 & 24.1 & 53.5  & 66.2 \\ 
6 & 7.68 & 24.2 & 53.7 & 66.1 \\
10 & 12.80 & 24.8 & 55.1 & 67.8 \\ 
\textbf{14} & 17.92 & \textbf{27.3} & \textbf{56.6} & \textbf{68.7}  \\ 
18 & 23.04  & 26.9 & \textbf{56.6} & 68.5\\ 
\bottomrule
\end{tabular}
\caption{\textbf{Effect of stride on MSR-VTT performance, which affects the temporal span of a single clip.} All models are trained with RGB-only, using K400 initialisation, 32 input frames and a batch size of $64$. At test time, we sample 4 equally spaced clips and average the similarity scores. Best performance is obtained with a stride of $14$.} 
\label{tab:clip-coverage}
\end{table}
\begin{table}[h]
\centering
\begin{tabular}{lccc}
\toprule
Init. & Modality & R@1 & R@10 \\
\midrule

Scratch & A & 19.1 & 64.7 \\ 
ImageNet21K~\cite{deng2009imagenet} & A  & 30.2 & 75.4 \\
VGGSound~\cite{chen2020vggsound} & A & \textbf{32.0} & \textbf{82.3} \\
\bottomrule
\end{tabular}
\caption{\textbf{Audio encoder initialisation on the AudioCaps dataset for text-audio retrieval.} }
\label{tab:audiocaps-ablation}
\end{table}
\section{Implementation Details} 
In this section we provide more details about the inputs to the MBT video encoder. RGB frames for all datasets are extracted at 25 fps. For MSR-VTT we sample $32$ RGB frames with stride $3$ frames, while for AudioCaps we sample $8$ RGB frames with a uniform stride of $56$ frames. Audio for all datasets is sampled at 16kHz and converted to mono channel. Following MBT, we extract log mel spectrograms with a frequency dimension of 128, 25ms Hamming window and hop length 10ms. This gives us an input of size $128 \times 100$ for 1 second of audio. We sample $8$ audio spectrograms for each video clip, and unlike MBT, we use a stride of $3$ between spectrograms to cover $24$ seconds of audio at a time. For the MSR-VTT data set examples missing audio, we feed in zeros as input.  
\section{Model architecture ablations}
In this section we provide ablations on the stride of frames used in the video encoder as well as the initialisation of the audio encoder for the AudioCaps dataset. 
\subsection{Clip coverage}
We use the stride of the sampled RGB frames to control the coverage of clips that are randomly sampled during training, and provide the results in Table \ref{tab:clip-coverage}. A randomly sampled 2 second clip from the video (stride=2) does much worse than using a stride of 14 (18s clip coverage). We find in general a greater clip coverage leads to better performance, indicating that the captions in MSR-VTT usually refer to concepts that either span the entire clip, or that may be missed by randomly sampling a 2s segment. This observation was also made by FIT~\cite{bain2021frozen}.  Note that the numbers here are lower than our best model, as we use a batch size of $64$ during training (compared to $256$ used for our best model). 

\subsection{Audio encoder initialisation}
We experiment with initialising the MBT backbone with ImageNet-21K and VGGSound weights, for the task of audio retrieval on the AudioCaps dataset. Results are in Table \ref{tab:audiocaps-ablation}. Unlike the video initialisation ablation in Table 1 of the main paper, we find that VGGSound initialisation provides a large improvement over Imagenet, and use this as a default for experiments on both the AudioCaps and Clotho datasets.


\end{document}